\documentclass[sigconf,
    ]{acmart}

\usepackage[nolist]{acronym}
\usepackage{booktabs}
\usepackage{tabu}
\usepackage{dcolumn} 
\usepackage{academicons}

\renewcommand\footnotetextcopyrightpermission[1]{} 

\acmConference[ACM e-Energy '23]{The 14th ACM International Conference on Future Energy Systems}{June 16 -- 23,  2023}{Orlando, Florida}
\begin{document}

\begin{acronym}
\acro{ANN}{Artificial Neural Network}
\acro{ARIMA}{Auto-Regressive Integrated Moving Average}
\acro{DSO}{Distribution System Operator}
\acro{kNN}{$k$ nearest Neighbor}
\acro{LSTM}{Long Short Term Memory}
\acro{ML}{Machine Learning}
\acro{MAE}{Mean Absolute Error}
\acro{MAPE}{Mean Absolute Percentage Error}
\acro{MRA}{Meta Regression Analysis}
\acroplural{MRA}[MRAs]{Meta-Regression Analyses}
\acro{MSE}{Mean Square Error}
\acro{NRMSE}{Normalized Root Mean Square Error}
\acro{NN}{Neuronal Network}
\acro{OLS}{Ordinary Least Squares}
\acro{RQ}{Research Question}
\acro{RNN}{Recurrent Neural Network}
\acro{RMSE}{Root Mean Square Error}
\acro{SMAPE}{Symmetric Mean Absolute Percentage Error}
\acro{SVM}{Support Vector Machine}
\acro{SVR}{Support Vector Regression}
\acro{WLS}{Weighted Least Squares}
\end{acronym}

\title{Meta-Regression Analysis of Errors in Short-Term Electricity Load Forecasting}


\author{Konstantin Hopf}
\affiliation{%
  \institution{University of Bamberg}
  \city{Bamberg}
  \country{Germany}}
\email{konstantin.hopf@uni-bamberg.de}
\orcid{0009-0006-5522-0416}

\author{Hannah Hartstang}
\affiliation{%
  \institution{University of Bamberg}
  \city{Bamberg}
  \country{Germany}}
\email{hannah.hartstang@gmx.de}
\orcid{0009-0006-5522-0416}

\author{Thorsten Staake}
\affiliation{%
  \institution{University of Bamberg}
  \city{Bamberg}
  \country{Germany}}
\email{thorsten.staake@uni-bamberg.de}
\orcid{0000-0003-1399-4676}


\begin{abstract}
  Forecasting electricity demand plays a critical role in ensuring reliable and cost-efficient operation of the electricity supply. With the global transition to distributed renewable energy sources and the electrification of heating and transportation, accurate load forecasts become even more important. While numerous empirical studies and a handful of review articles exist, there is surprisingly little quantitative analysis of the literature, most notably none that identifies the impact of factors on forecasting performance across the entirety of empirical studies. In this article, we therefore present a Meta-Regression Analysis (MRA) that examines factors that influence the accuracy of short-term electricity load forecasts. We use data from 421 forecast models published in 59 studies. 
  While the grid level (esp. individual vs. aggregated vs. system), the forecast granularity, and the algorithms used seem to have a significant impact on the MAPE, bibliometric data, dataset sizes, and prediction horizon show no significant effect. We found the LSTM approach and a combination of neural networks with other approaches to be the best forecasting methods. 
  The results help practitioners and researchers to make meaningful model choices. Yet, this paper calls for further MRA in the field of load forecasting to close the blind spots in research and practice of load forecasting.
\end{abstract}


%



\maketitle

\textit{\copyright Hopf, Hartstang, Staake (2023). This is the author's version of the work. It is posted here for your personal use. Not for redistribution. The definitive version was published in the Proceedings of the 14th ACM International Conference on Future Energy Systems (e-Energy '23), June 20--23, 2023, Orlando, FL, USA, \url{https://doi.org/10.1145/3599733.3600248}.}

\section{Introduction}

Accurate forecasting of electricity demand is an important success factor for utilities, and there is reason to believe that such forecasts will become even more important in the future:
The electrification of residential heating and the adoption of electric vehicles will increase both, volatility of demand and utilization of the distribution grid \cite{Chapaloglou.2019}. As a result, the safety margins of existing assets will decrease while energy costs at peak load times will rise. 
Load control---both centralized, e.g., by grid operators, and decentralized by local agents---will benefit from accurate demand forecasts, as will utilities’ attempts to schedule production and hedge demand through forward contracts.

As a consequence, there is an extensive literature on methods and models for electric load forecasting \cite{Hong.2020}, which has been summarized in recent review papers \cite[e.g.,][]{Haben.2021, Sun.2020, Nti.2020}. The academic discourse is fueled by the continued global deployment of advanced metering infrastructures that makes an increasing amount of consumption data available at higher temporal resolution. Thus, such data have attracted significant research interest to investigate the use of smart meter data for load forecasting and to enable electricity forecasting at different grid levels and for different time scales \cite{Wang.2018}.

The existing literature reviews on short-term electricity load forecasting provide good overviews of the large number of empirical studies. They show which computational methods are used to forecast future electricity demand for different forecast horizons and grid levels. The reviews also identify blind spots in existing research and outline research agendas. Yet, the existing review studies are primarily qualitative.

Quantitative meta-reviews, by contrast, aim to build a mean effect size from comparable, independent individual studies. Thus, quantitative meta-reviews allow for more reliable results than single empirical studies \cite{Nelson.2009}. They also enable the identification of parameters that explain variances and heterogeneity of effect sizes in study results when sufficient data are available \cite{stanley_meta-regression_2012, cooper_handbook_2009}. 
Such quantitative meta-reviews are particularly helpful for practitioners who strive to operationalize forecasts for specific situations and want to draw on evidence from a complete research field \cite{cooper_handbook_2009}. 
They also help research to judge the robustness of existing approaches, recognize patterns within working solutions, and identify outliers that might be especially promising or questionable. 

Our study seeks to extend the previous reviews with an inductive statistical analysis, as we carry out a \ac{MRA} for short-term electricity load forecasts. By short-term time horizon, we mean load forecasts with up to one week ahead \cite{Haben.2021, Bourdeau.2019, Nti.2020}. 
We thereby aim to explain factors---across a large sample of individual studies examining electric load forecasting---that lead to high or low quality short-term electric load forecasts.

Our article starts with an overview of recent electric load forecasting review studies, describes the method, and our analytical results. We conclude with an interpretation and outline future directions for research and practice.

\section{Background}

The field of electricity load forecasting is comprehensive and stays in connection to other fields of energy forecasting \cite{Hong.2020}. We found several review articles that were published in the last five years (see \autoref{tab:review_studies}) and provide a summarized overview of the field. 
Similar to trends that Hong et al. \cite{Hong.2020} identify for the broader field of energy forecasting, the literature on electricity load forecasting heavily uses developments in the field of \ac{ML}. 
Another observation that holds for the fields of energy and electricity load forecasting alike is that studies primarily focus on forecasts at the system level or the transmission grid and (through the proliferation of smart meter data in recent years) also on the household level \cite{Haben.2021, Hong.2020}. Other grid levels have not been in the focus yet. 

\begin{table}[h]
    \centering
    \footnotesize
    \caption{Literature reviews on electricity forecasting in the last five years}
    \label{tab:review_studies}
    \begin{tabu}{llXll}
\toprule
Ref. & Year & Journal & Dep. variable & Focus \\
\midrule
\cite{Haben.2021}*  & 2021 & Appl. En.              & load& low voltage grid\\
\cite{Aslam.2021}   & 2021 & Ren. and Sust. En. Rev.& load + prod.& no restriction\\
\cite{Walther.2021} & 2021 & Energies               & load& manufacturing\\
\cite{Nti.2020}*    & 2020 & J. El. Sys. and IT     & load& no restriction\\
\cite{Sun.2020}*    & 2020 & En. and Buildings      & load& buildings\\
\cite{Vivas.2020}   & 2020 & Entropy                & load + prod.&no restriction\\
\cite{Devaraj.2021}	& 2020 & Intl. J. of En. Res.   & load + prod.&no restriction\\
\cite{Bourdeau.2019} &2019 & Sust. Cities and Society & load + prod.& buildings\\
\cite{Wang.2018}*   & 2018 & IEEE Tran. on Smart Grid & load& no restriction\\
\midrule

    \end{tabu}
\end{table}

The review articles describe the landscape of electricity load forecasting, list algorithmic approaches, datasets, various forecasting horizons, and grid levels. They also problematize implicit field assumptions, point out limitations in the field, and identify future research directions. Nevertheless, the review articles we found are primarily qualitative summaries. If the reviews include quantitative analyses, the evaluations of forecasting models focus on bibliometrics (e.g., publication date, journal) or analyze the prediction models in a descriptive way (e.g., frequency of algorithm categories). Only Vivas et al. \cite{Vivas.2020} investigate the relationship of broad algorithm classes (statistical vs. machine learning vs. hybrid) and data granularity on model performance, but are not examining influence factors on prediction performance with inductive statistics.

Thus, for practitioners that want to operationalize forecasts, it is difficult to decide which algorithmic approach is suitable for a given application when such aggregated knowledge does not exist.

\section{Method}
The approach of \ac{MRA} goes back to Glass \cite{Glass.1976} who proposed the method in 1976 as "the statistical analysis of a large collection of analysis results from individual studies for the purpose of integrating the findings" \cite[][p. 3]{Glass.1976}. 
\acp{MRA} find wide application in many fields such as medicine, psychology, and economics \cite{Nelson.2009, stanley_meta-regression_2012}. In research related to the energy domain \acp{MRA} exist, for example, on water demand \cite{Sebri.2016} and energy prices \cite{Gurtler.2018}.

\acp{MRA} typically follow the following three steps  \cite{Stanley.2013}: First, define the research question and the effect size (as the core criterion of interest), second, literature search and coding, and third, meta-regression modeling. Our description below follows these three steps.

\subsection{Review focus and effect size}
Our focus lies on the predictive quality of short-term electricity forecasts (up to week-ahead) using point-estimates. To evaluate such forecasts, several performance metrics exist \cite{james_introduction_2013}. Absolute error metrics, like \ac{MAE} or \ac{RMSE} are not helpful as they are scaled in the dimension of the input data, thus, do not allow comparison across studies. 

An error measure that is scale-independent and allows comparisons across studies is the \ac{MAPE} metric. As an alternative, the \ac{NRMSE} could also be used, but we found that \ac{MAPE} is more frequently reported (77.2\% of studies in our sample reported \ac{MAPE} and only 12.7\% \ac{NRMSE}). Thus, we selected \ac{MAPE} as the effect size.

\subsection{Literature selection}

As \acp{MRA} aim to provide a comprehensive overview to a field, the selection of the sample of empirical studies included into the analysis is crucial and should be consciously made \cite{stanley_meta-regression_2012}. We decided to use empirical studies that were mentioned in four review articles on electricity load forecasting that appeared in the last five years. All review articles report a systematic review process with clear selection criteria \cite{Higgins.2022, Chalmers.2002}. Each review article covers a slightly different review focus, which increases the breath of our sample. 

In detail, we use the review study by Nti et al. \cite{Nti.2020}, which focuses on electricity load forecasting in general, the review by Haben et al. \cite{Haben.2021}, which focuses on short-term electricity forecasting in the low-voltage grid, the review of Sun et al. \cite{Sun.2020}, which focuses on forecasts electricity use of buildings, and the review by Wang et al. \cite{Wang.2018}, which focuses on the use of smart meter data. We mark the used studies with an asterisk in \autoref{tab:review_studies}.  

\begin{figure}
    \centering
    \includegraphics[width=\linewidth]{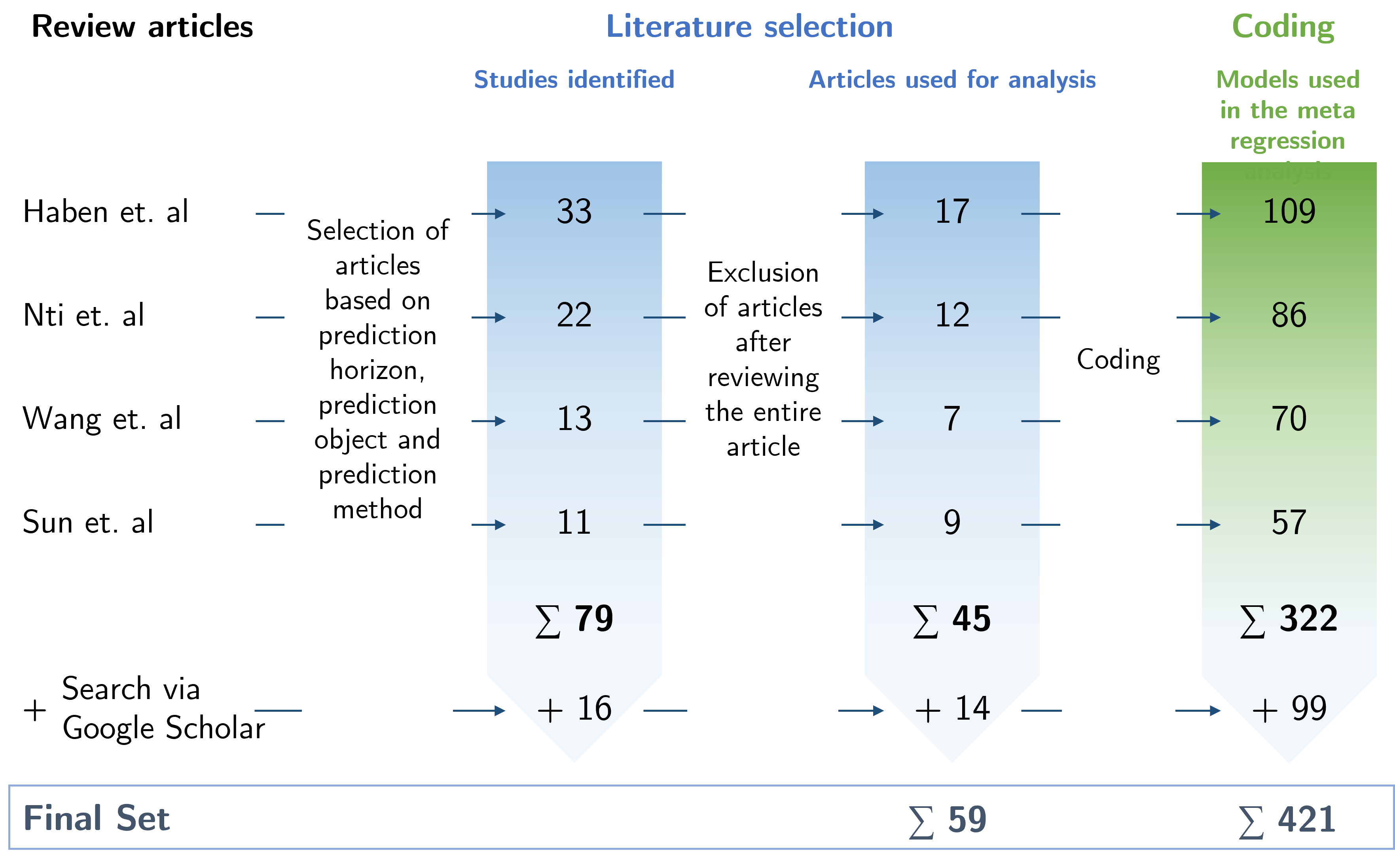}
    \caption{Literature selection and coding process}
    \label{fig:lit_selection_coding}
\end{figure}

Our selection of empirical studies from these review papers followed three steps, as we illustrate in \autoref{fig:lit_selection_coding}. 
We first screened each review paper for empirical papers that investigated a short-term prediction horizon (up to week-ahead), used electric load as a dependent variable and obtained point estimates (we excluded studies on probabilistic forecasting). We made this initial selection of articles based on the information presented in the review article. This initial screening led us to 79 references. 

Second, we had to exclude studies after our in-depth reading with the following reasons: As we use \ac{MAPE} as an effect size metric, we must exclude articles that do not report \ac{MAPE} values or do not clearly document the actual numbers (e.g., that just show bars in graphs without actual numbers). In addition, we excluded review studies or secondary studies that just compared results of other primary studies. Furthermore, we excluded studies that did not allow any conclusions about the sample size of the data used for training and evaluation. In total, 45 articles remained after completing this second article screening step.

As a third step, we conduced a limited structured search of articles via Google Scholar, in order to complete the sample for our \ac{MRA}. We used the search terms "short-term", "forecast*", "MAPE", "energy", "load", and "demand" and reviewed the hits on the first pages according to the first and second step described above. This search yielded to additional 16 articles, of which we included 14 in our analysis, leading us to a sample of 59 articles.

\subsection{Coding}

\begin{table}[b]
    \centering
    \small
    \caption{Numeric variables}
    \label{tab:variables_numeric}
    \begin{tabu}{lXrr}
    \toprule
        Variable & Description & Mean & Std. Dev.\\
        \midrule
        MAPE & Percent & 16.07 & 23.92 \\
        Horizon\_Num	& In half-hour steps & 	49.7	& 60.1	\\
        Granularity\_Num	& In half-hourly values & 7.83 &	15.76 \\
        Year & Year of publication & 2017.39	& 2.64  \\
        N\_days & Days for model training and test 	& 460.2 	& 486.7	\\
        N\_obs & No. of observation points  & 334 &	3{,}948 \\
    \bottomrule
    \end{tabu}
\end{table}

During our in-depth reading of the articles, we extracted all models that the studies report together with all relevant information for our \ac{MRA}. For this, we use a systematic coding procedure \cite{krippendorff_content_2018}, following the coding guide that we describe in the remaining section. We list the resulting variables in \autoref{tab:variables_numeric} and in \autoref{tab:variables_cat}. 

\textbf{Effect size and sample sizes}:
We coded the \ac{MAPE} values in percent, the number of observation points as integer, and transformed the timespans of the data reported in the papers as the number of days.

\textbf{Forecast horizon}:
We coded the forecast horizon $h$ in a categorical and in a numeric variable. For the categorical, we differentiate between four ordered classes {$h \leq 1hourahead$, $1hourahead < h < 1dayahead$ , $h=dayahead$ , $h>dayahead$}. For the numeric coding, we gathered the number of half-hour steps.

\textbf{Forecast granularity}:
We also coded the forecast granularity, which is the resolution in which the forecast is computed. To harmonize this across the different studies, we express the granularity relative to a half-hour step, i.e., hourly granularity would be 2 and quarterly would be 0.5. 

\textbf{Model category}:
Finally, we classified the models in the studies into one of 17 categories of algorithmic approaches. For this, we created a classification scheme based on schemata used in the considered review papers \cite{Haben.2021, Nti.2020, Sun.2020, Vivas.2020} but also others \cite{Hammad.2020, Debnath.2018}. A common division of forecasting models makes a distinction between statistical, \ac{ML}, and hybrid models. We further differentiate these three groups into different algorithmic approaches.

\begin{table}[b]
    \centering
    \small
    \caption{Categorical variables}
    \label{tab:variables_cat}
    \begin{tabu}{lXr}
    \toprule
        Variable & Categories & No. Models\\
        \midrule
        Level 
            & Individual (household, building, ...) &			146 (34.7\%)\\
            & Aggregated (sum of multiple entities) & 			105 (24.9\%)\\
            & Substation (transformer, grid zone, ...) & 23 (5.4\%)\\
            & System                                          & 147 (34.9\%)\\
        \midrule
        Horizon\_Cat 
            & $h \leq 1hourahead$	&	64 (15.2\%) \\
            & $1hourahead < h < 1dayahead$	  &	29 (6.9\%)\\
            & $h=dayahead$ 	 &		308 (73.2\%)\\
            &  $h>dayahead$		& 20 (4.7\%)\\
        \midrule
        Model\_Cat
            & Stat: Time Series & 52 (12.3\%)\\
            & Stat: Regression & 42 (10.0\%)\\
            & Stat: Exponential Smoothing & 16 (3.8\%)\\
            & ML: Shallow\_NN	& 82 (19.5\%)\\
            & ML: Deep\_NN & 50 (11.9\%)\\
            & ML: RNN &13 (3.1\%)\\
            & ML: LSTM &29 (6.9\%)\\
            & ML: Boltzmann & 3 (0.7\%)\\
            & ML: SVM\_SVR & 50 (11.9\%)\\
            & ML: kNN &7 (1.7\%)\\
            & ML: Fuzzy\_Logic	 &14 (3.3\%)\\
            & ML: Ensemble and other & 15 (3.6\%) \\
            & Hybrid\_NN	& 	22 (5.2\%)\\
            & Hybrid\_various	& 2 (0.47\%)\\
            & Benchmark & 24 (5.7\%)\\

        \midrule
        Study\_ID & \textit{Unique per study} & 421 (100\%)\\
        \bottomrule
    \end{tabu}
\end{table}

Among \textit{statistical methods}, we differentiate between \textit{time series} (including AR(X), NARX, MA, ARMA(X), ARIMA(X), SARIMA(X), state space, and spatio-temporal models), \textit{regression} (including linear and multilinear regressions) and \textit{exponential smoothing} methods (e.g., Holt-Winters models). 

In the category of \textit{\ac{ML} models}, \acp{NN} are the most frequent algorithm category. We split this category into \textit{Shallow\_NN} and \textit{Deep\_NN}, where \textit{Shallow\_NN} are those having only one hidden layer and \textit{Deep\_NN} are those having more than one hidden layer. To keep the number of categories manageable, we 
just differentiate between network architectures, in particular, architectures for time-series data, i.e., \acp{RNN} and \acp{LSTM} (which is an advancement of the \ac{RNN} approach). 
We combined \ac{SVM} and \ac{SVR} into one category. 
Other machine learning models include \textit{Boltzmann} machine models, \textit{\ac{kNN}}, \textit{fuzzy logic}, and \textit{ensemble and other} models (e.g., autoencoders, genetic algorithms, gradient boosting, random forest,  and decision tree algorithms).

Several studies propose \textit{hybrid models} by combining an algorithm from one of the considered classes with another (e.g., a linear regression model). We consider two classes of hybrid models, those with \acp{NN} (\textit{Hybrid\_NN}) and those with others (\textit{Hybrid\_various}).  

Finally, multiple studies use \textit{benchmark estimators}, which do not transform the data in a sophisticated way. These simple models are grouped in their own class.

\begin{figure}[h]
    \centering
    \includegraphics[width=\linewidth]
        {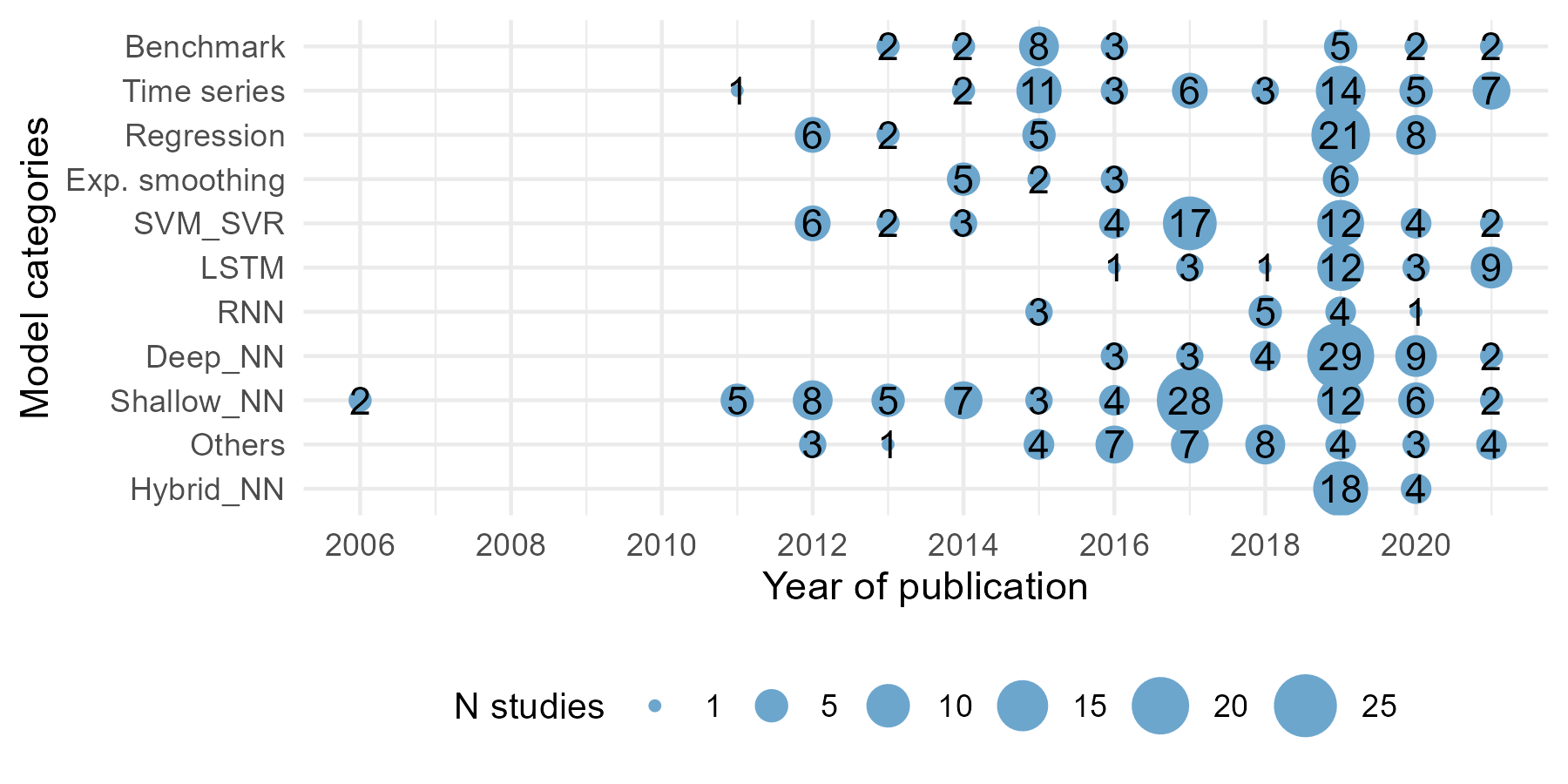}
    \vspace*{-1em}
    \caption{Illustration of the use of algorithmic categories in the studies over time}
    \label{fig:categories_time}
\end{figure}

To get a better understanding of the use of the different model categories in the studies, \autoref{fig:categories_time} shows in the number of studies that use the different algorithm classes over time. Looking at the \ac{NN}-based approaches, we notice that before 2018 mainly \textit{Shallow\_NN} were used. After 2018, the models with multiple hidden layers dominated. This indicates an evolution in research on \acp{NN}. \acp{LSTM} are gaining  momentum over time and are the most common approach used in the studies published in 2021. Time series and benchmark models are used fairly regularly over the years and most often serve as reference models in the studies to compare. We also find that some approaches are used only very rarely, for example, fuzzy logic, ensemble methods, kNNs, Boltzmann Machine, Genetic Algorithms, and Autoencoders. We therefore group these infrequent model categories together for the following analyses.

\subsection{Statistical analysis}

For the statistical analysis of our \ac{MRA}, we rely on \ac{OLS} and \ac{WLS} regression. Guidelines for conducting \acp{MRA} point to the problems of heteroscedasticity, reliability and interdependence between examined factors \cite{Nelson.2009, Sebri.2016}, which we address through the following approaches. 

First, to mitigate the problem of heteroscedasticity, we use robust standard errors \cite{white_heteroskedasticity-consistent_1980, zeileis_econometric_2004}. Second, to include an estimate of the study reliability, we weight the effect sizes of the single empirical studies. Meta-analyses frequently use the variance of the effect sizes as a weighting factor, given that effect sizes with smaller variances are considered as more reliable and should be weighted more heavily in a \ac{MRA}. In the field of electricity forecasting, the variance of error metrics are, however, usually not reported. Therefore, we estimate the reliability of a study using the sample size \cite{Nelson.2009}. As we have time-series data available, and ideally time series data from multiple observation points, we compute
\begin{equation}
\label{eq:samplesize}
SampleSize = N\_days * N\_obs
\end{equation}
\noindent
As some studies have very large samples, we use the logarithm to lower the influence of very large sample sizes and scale the weighting factors using max-normalization.

Third, we evaluate interdependence between examined influence factors and the effect size. One reason can be that multiple primary studies use the same data set, which is party the case for electricity forecasting \cite{Haben.2021}. Another reason is that multiple effect sizes may be reported from a single study. Observable common effects, such as the common data set, can be accounted for using regressors. To account for study-specific influences, we use fixed effects models in that we estimate a regression intercept per study \cite{Nelson.2009, Borenstein.2010}.

\section{Results}
\subsection{Study-specific parameters}

For the models that examine study-specific characteristics, we computed simple linear models using \ac{OLS} to test the correlation between \ac{MAPE} and one variable as regressor each. First, we could not find an influence of the year of publication on the error values  ($R^2=0.01$, $F(421) = 2.36$, $p = .1612$). 
Second, we tested the influence of the size of the data set, that is, the observation period in days and the number of observation points (e.g., meters or households). While the regression models found statistically significant effects of the size of the data, the effect sizes are very small and the variances explained by the study parameters are low for the observation period ($R^2 = .000$, $F(421) = 0.39$, $p < .001$) and the number of observation points ($R^2 = .008$, $F(421) = 3.48$, $p < .05$). 

\subsection{Grid levels}

As a second analysis, we examined the influence of the grid level on the forecasting error. We encoded the grid level as dummy variables, $d_{ind}=1$ if the forecast was made on an individual level (e.g., households, buildings), $d_{sub}=1$ if the forecast was obtained for the substation-level, and $d_{aggr}=1$ if the forecast was made by aggregating time-series from lower grid levels. The case that the forecast targeted the system level is represented as response state (i.e., all dummy variables are zero) because this is the most frequent grid level across all studies. We use a fixed-effects model considering an intercept for each study (represented by the study $ID_i, i\in{1, ..., 59}$) with the coefficient $\beta_{0i}$. We estimated the model using a \ac{WLS} estimation using $log(SampleSize)$ as a weighting factor (see \autoref{eq:samplesize}) and used robust standard errors to address heterogeneity \cite{white_heteroskedasticity-consistent_1980, zeileis_econometric_2004} with the following model specification:
\begin{equation}
y_i = \beta_{0i} * ID_i + \beta_1 * d_{ind} + \beta_2 * d_{sub} + 
\beta_3 * d_{aggr} + \epsilon_i    
\end{equation}

\begin{table}[b]
    \caption{MAPE estimates for different grid levels, using the individual household as a baseline (model 1)}
    \label{tab:results_levels}

\begin{tabular}{l D{)}{)}{9)3}}
\toprule
 & \multicolumn{1}{c}{Model 1} \\
\midrule
Individual & 25.16 \; (4.16)^{***} \\
Aggregated  & -0.62 \; (3.15)       \\
Substation  & 0.18 \; (2.12)        \\
\midrule
R$^2$       & 0.72                  \\
Adj. R$^2$  & 0.67                  \\
Num. obs.   & 421                   \\
F statistic & 14.79                 \\
\bottomrule
\multicolumn{2}{l}{\tiny{$^{***}p<0.001$; $^{**}p<0.01$; $^{*}p<0.05$}}
\end{tabular}

\end{table}

\autoref{tab:results_levels} shows the \ac{WLS} model estimates. Due to the study-specific intercept, the proportion of explained variance is quite high (this holds also true for the following models). Accordingly, we focus our interpretation on the main effects and their significant difference from zero. 

\begin{figure}[h]
    \centering
    \includegraphics[width=\linewidth]{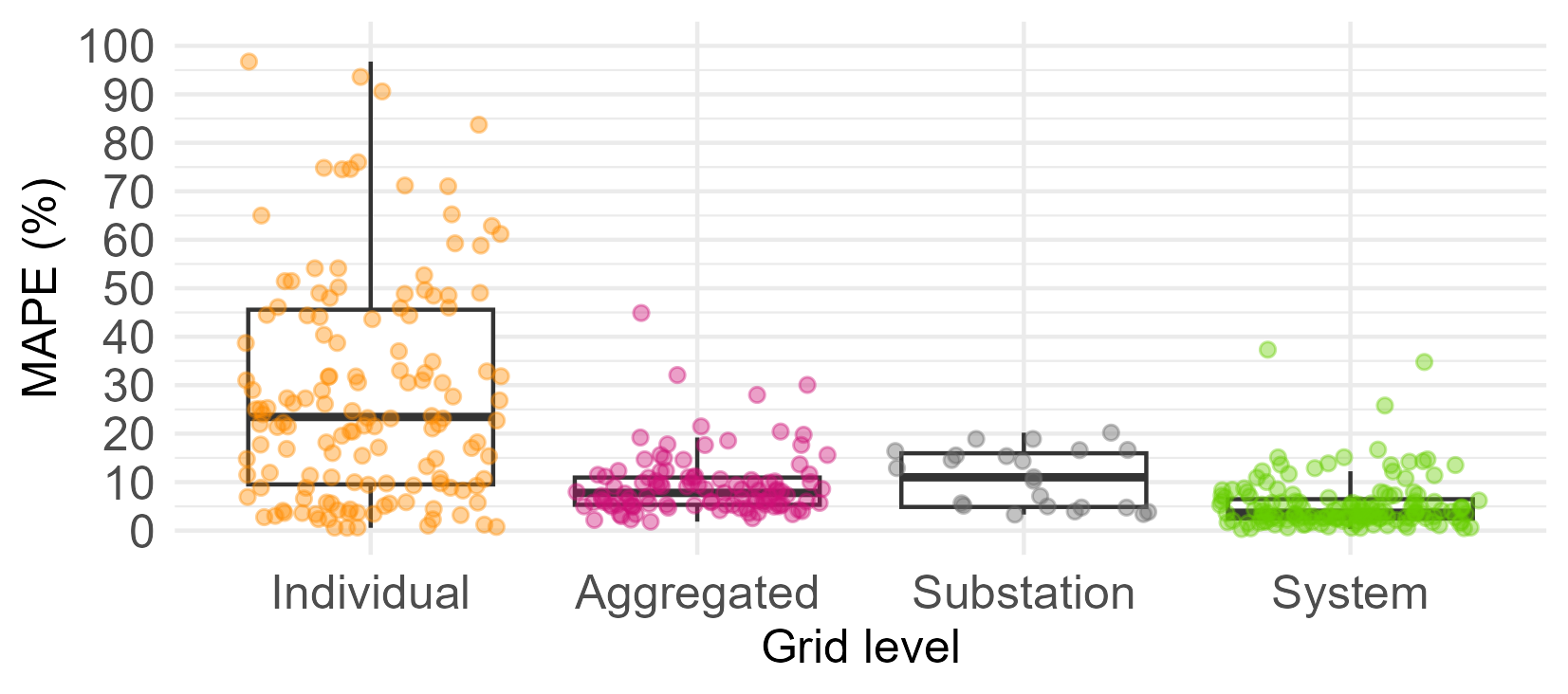}
    \vspace*{-1.5em}
    \caption{Combined scatter- and boxplot for MAPE values across different grid levels}
    \label{fig:mape_gridlevels}
\end{figure}

As the descriptive plot in \autoref{fig:mape_gridlevels} shows, the studies with individ\-ual-level prediction have a large variance and report significantly higher error values than the models with other grid levels. Compared to prediction for a single time-series of a complete power system (reference category), our model estimates that the forecast errors on individual level are 25.16 percentage points higher. If predictions are made with aggregated data, this seems to lead to better predictions (in 0.62 percentage points lower error), yet the difference is not statistically significant. The lacking significant level for forecasts at the aggregated and the secondary grid level (i.e., substation) might be due to the small number of studies that consider this level of forecast.

Reasons for the high relative errors in individual-level studies may be that individual load curves might be harder to predict than aggregated ones. Another explanation is that individual load curves often have times with small load, which increases the \ac{MAPE}. To investigate this issue further, other error metrics that give less weight to small consumption values may need to be included in the analysis.

\subsection{Time horizon}

Similar to the previous analysis, we tested if the forecasting horizon has an impact on the errors. 
Only few models had a horizon between hour- and day-ahead (29 models), and even fewer have a forecasting horizon of more than day-ahead (20 models). Thus, we excluded these models for this analysis. For the remaining data (372 models), we encoded the day-ahead as $d_{dayahead}=1$ and the hour-ahead forecasts as $d_{dayahead}=0$ and estimate the following regression model: 
\begin{equation}
y_i = \beta_{0i} * ID_i + \beta_1 * d_{dayahead} \epsilon_i    
\end{equation}
Given that the dummy encoding omits ranking information and we left out several models, we used an alternative model formulation with a metric variable. We defined a variable that expresses the forecasting horizon in the number of 30-minute time steps (i.e., an hourly forecast has $horizon\_timesteps=2$ and a 24h forecast $horizon\_timesteps = 48$). This approach follows earlier meta-reviews in the field of electricity \cite{Sebri.2016} and traffic forecasting \cite{Varghese.2020}.
\begin{equation}
y_i = \beta_{0i} * ID_i + \beta_1 * horizon\_timesteps + \epsilon_i    
\end{equation}
\autoref{tab:results_horizon} shows the \ac{WLS} model estimates. From both models, it appears that there is no significant influence of the forecast horizon on model quality. This may be because the effect is confounded by algorithm choice or practical relevance over time. The relatively high $R^2$ values result from the study-specific intercept in our fixed-effects model.

\begin{table}[h]
    \caption{MAPE estimates for different time horizons, using dummy (model 2) and using numeric encoding (model 3)}
    \label{tab:results_horizon}

\begin{tabular}{l D{)}{)}{8)0} D{)}{)}{8)0}}
\toprule
 & \multicolumn{1}{c}{Model 2} & \multicolumn{1}{c}{Model 3} \\
\midrule
Dayahead         & 0.63 \; (5.36) &                \\
Horizon\_Num &                & 0.01 \; (0.05) \\
\midrule
R$^2$                & 0.67           & 0.68           \\
Adj. R$^2$           & 0.61           & 0.63           \\
Num. obs.            & 372            & 421            \\
F statistic          & 10.91          & 12.73          \\
\bottomrule
\multicolumn{3}{l}{\tiny{$^{***}p<0.001$; $^{**}p<0.01$; $^{*}p<0.05$}}
\end{tabular}

\end{table}

\subsection{Forecast granularity}

As a fourth analysis, we examine the influence of the forecast granularity on the size of the error, using the model: 
\begin{equation}
y_i = \beta_{0i} * ID_i + \beta_1 * granularity + \epsilon_i    
\end{equation}
The regression results in \autoref{tab:results_granularity} show that granularity has a negative and significant effect on the magnitude of the prediction error ($R^2 = .68$, $F(60, 361) = 12.9$, $p < .001$). This means that larger time steps of the forecasts lead to lower relative prediction errors. For every half hour that the forecast granularity increases, the error decreases by 0.39 percentage points.

\begin{table}[h]
    \caption{MAPE estimates for different data granularities}
    \label{tab:results_granularity}

\begin{tabular}{l D{)}{)}{9)2}}
\toprule
 & \multicolumn{1}{c}{Model 4} \\
\midrule
Granularity\_Num & -0.39 \; (0.14)^{**} \\
\midrule
R$^2$                            & 0.68                 \\
Adj. R$^2$                       & 0.63                 \\
Num. obs.                        & 421                  \\
F statistic                      & 12.90                \\
\bottomrule
\multicolumn{2}{l}{\tiny{$^{***}p<0.001$; $^{**}p<0.01$; $^{*}p<0.05$}}
\end{tabular}
\end{table}
 
\begin{table*}[h]
    \caption{MAPE results for algorithm category (overall and across different grid levels)}
    \label{tab:results_algorithms_hierarchical}

\begin{tabular}{l D{)}{)}{10)5} D{)}{)}{11)1} D{)}{)}{9)2} D{)}{)}{9)3}}
\toprule
 & \multicolumn{1}{c}{Overall} & \multicolumn{1}{c}{Individual} & \multicolumn{1}{c}{Aggregated} & \multicolumn{1}{c}{System} \\
\midrule
Shallow\_NN          & 1.51 \; (4.28)          & 3.27 \; (11.41)       & -0.76 \; (2.53)      & -3.57 \; (1.27)^{**}  \\
Deep\_NN             & -8.74 \; (4.48)^{\cdot} & -15.38 \; (13.29)     & -8.26 \; (2.55)^{**} & 4.80 \; (1.24)^{***}  \\
Time\_Series           & -0.56 \; (4.25)         & -2.69 \; (12.66)      & -2.96 \; (2.54)      & -2.78 \; (1.20)^{*}   \\
RNN          & -7.16 \; (6.24)         & -16.83 \; (15.89)     & -7.46 \; (3.20)^{*}  & -1.16 \; (2.17)       \\
LSTM                   & -15.85 \; (5.26)^{**}   & -27.54 \; (11.61)^{*} & -8.66 \; (2.96)^{**} &                       \\
SVM\_SVR               & -3.69 \; (4.98)         & -1.78 \; (13.26)      & -3.60 \; (2.97)      & -4.86 \; (1.30)^{***} \\
Benchmark              & -3.81 \; (5.37)         & -19.94 \; (14.21)     & 1.52 \; (2.82)       & 1.40 \; (2.22)        \\
Exponential\_Smoothing & -1.34 \; (5.54)         & -6.61 \; (15.86)      & -3.68 \; (2.83)      & 5.18 \; (2.05)^{*}    \\
Hybrid\_NN             & -15.65 \; (5.89)^{**}   & -34.89 \; (15.77)^{*} & -8.37 \; (2.94)^{**} & -1.26 \; (5.00)       \\
Others             & -0.57 \; (4.81)         & -0.37 \; (11.74)      & -7.53 \; (2.82)^{**} & -4.27 \; (1.49)^{**}  \\
\midrule
R$^2$                  & 0.69                    & 0.77                  & 0.87                 & 0.88                  \\
Adj. R$^2$             & 0.63                    & 0.69                  & 0.82                 & 0.85                  \\
Num. obs.              & 421                     & 146                   & 105                  & 147                   \\
F statistic            & 11.56                   & 9.81                  & 18.64                & 29.10                 \\
\bottomrule
\multicolumn{5}{l}{\tiny{$^{***}p<0.001$; $^{**}p<0.01$; $^{*}p<0.05$; $^{\cdot}p<0.1$}}
\end{tabular}

\end{table*}

\subsection{Algorithm category}

Finally, we investigate the influence of the model category on the forecasting error. As algorithmic innovations aim to improve forecasts, we expect a strong influence of the model category on the forecast errors  \cite{Vivas.2020, Kuster.2017}. 

We consider the algorithm class as a dummy-encoded variable $Model\_Category_j$ for each model category, being 1 if the class is used, 0 otherwise. We use \textit{regression models} as the response category (because they have the worst forecasting performance in the studied models), meaning that all $Model\_Category_j = 0$. Given that several model categories are very infrequent, we used ten model categories as shown in \autoref{tab:results_algorithms_hierarchical}, thus $1\leq j \leq 10$. 
\begin{equation}
y_i = \beta_{0i} * ID_i + \beta_j * Model\_Category_{ij} + \epsilon_i    
\end{equation}

The results of the regression analysis in \autoref{tab:results_algorithms_hierarchical} (\textit{Overall}) show that all model categories except the \textit{Shallow\_NNs} seem to produce better results than the \textit{regression} (which is the response category), even the benchmark models. Yet, only the coefficients of the model categories \textit{Deep\_NN}, \textit{\ac{LSTM}}, and \textit{Hybrid\_NN} have a significant effect on the magnitude of the forecast error overall. 

In our first analysis, we have found that the grid level has a significant influence on the model results. Therefore, we also analyzed the model categories with data subsets of studies focusing on an individual, aggregated, and system level (the substation category was too infrequent that we could compute the model). For the system level, there is only one study in our sample using the \textit{\acp{LSTM}} approach, thus, our model cannot estimate a coefficient due to perfect collinearity in this case. 

We see that \textit{\acp{LSTM}} and \textit{Hybrid\_NNs} show the lowest prediction errors and also have significant effects in the \textit{Individual} and \textit{Aggregated} subsample. Thus, we conclude that these two approaches lead to the best forecasting results in the sample of analyzed studies for the two grid levels.

For the grid level, the performance figures are quite different, suggesting that \textit{SVM\_SVR}, \textit{Shallow\_NN} and \textit{Other} approaches lead to lower results. Yet, the otherwise strong category of \textit{\ac{LSTM}} has been left out of the calculation because of just a single study. Future research should, thus, investigate the performance of algorithm categories on a system level including further and, in particular, more recent studies.

\section{Discussion}

The in-depth analysis of error metrics using a \ac{MRA} helps to identify parameters that explain variances and heterogeneity of effect sizes in a large number of empirical studies \cite{Nelson.2009, stanley_meta-regression_2012, cooper_handbook_2009}. For the field of short-term electricity load forecasting, \acp{MRA} can help to identify parameters, algorithms, and situations that foster smaller forecast errors.  

Our analysis showed, for example, that the grid level (esp. individual vs. aggregated vs. system), the forecast granularity, and the algorithms used have a significant impact on the \ac{MAPE}. We also found that the \ac{LSTM} approach and a combination of \acp{NN} with other methods were the best forecasting methods for the individual and aggregated forecasts. For system level forecasts, \ac{SVM}, \ac{SVR}, Shallow \acp{NN}, and other approaches seem to perform best in our sample. In contrast, the year of publication, dataset size, and prediction horizon had no significant effect on prediction performance in our sample of studies.

The results help practitioners to operationalize forecasting models for specific applications, drawing on the aggregated findings of 59 empirical studies. For researchers, the results help to assess the robustness of approaches they suggest, identify patterns and blind spots in the variety of existing solutions, and identify outliers that may be particularly promising or questionable.

\subsection{Future work}
The analysis we presented in this paper is promising and should be a call to the load forecasting research field to look more closely at \acp{MRA}. Many aspects could not be addressed in this study and thus require follow-up investigations. 

First, the sample of studies would need to be expanded to include older studies to better reflect trends over time in this area. Beyond that, a broader sample would strengthen the analysis. Extending the sample would also allow more detailed insights in subgroup analyses, for example, if some algorithms are better-suited for certain grid levels or data sources than others.

Second, the field of load forecasting is constantly evolving and new approaches, such as the Transformers architecture \cite{lim_temporal_2021, vaswani_attention_2017, giacomazzi_short-term_2023}, could not be included in our evaluation so far. 

Third, we also did not control for the datasets used. Even though many studies use datasets that are not public, there are some datasets that are used very often \cite{Haben.2021}. This may bias the analysis. An in-depth analysis in terms of datasets (also the statistical properties of the datasets used in the empirical studies) could provide further insight into, for example, how larger training datasets and high-resolution data have an impact on predictive performance. 

Fourth, an investigation of the influence of different data sources and features used (e.g., weather, geospatial information), also with a focus on open data \cite{hopf_mining_2018}, on forecast performance would also be an exciting extension. 

Finally, future research should also apply \acp{MRA} to probabilistic load forecasts. We could not include such studies because probabilistic load forecasts are evaluated with other error metrics and thus a different dependent variable would be necessary.

\subsection{Limitations}
A limitation of our analysis is the use of the \ac{MAPE} metric, which is highly dependent on actual consumption in the evaluation. We would have liked to use a more reliable quality metric, such as \ac{NRMSE} but other metrics that allow comparison between studies are rarely reported. Future research in the prediction literature should therefore report more quality metrics that allow quantitative comparative analysis.

\section{Conclusion}

The research literature on short-term electricity load forecasting is extensive, and previous survey articles summarize the field descriptively. To our knowledge, our analysis is the first \ac{MRA} to quantitatively examine factors influencing forecast quality. 

To close this gap, we have analyzed the prediction errors of 421 models published in 59 studies that were mentioned in recent review articles in the field of short-term electricity load forecasting. Our statistical \ac{MRA} could find statistically significant influences of (i) the grid level (individual, aggregated, and system), (ii) the forecast granularity, and (iii) the algorithms used (particularly good approaches are \ac{LSTM} and Hybrid\_NN on the individual and the aggregated level, while \ac{SVM}, shallow \ac{NN}, and other \ac{ML} approaches perform best for grid-level forecasts) on the \ac{MAPE} reported in these studies. We did not find an influence of the study characteristics (year of publication, dataset size) or the time horizon of the forecast on the \ac{MAPE}. 

Although short-term load forecasting offers powerful tools with acceptable forecast error metrics, the development of new forecasting methods remains a major challenge. In the future, the influences of the energy transition will have a greater impact on all energy consumption sectors. For example, heat pumps and electric vehicles will proliferate, causing additional loads. Weather influences on electricity consumption will change, caused by environmental change and increase power-to-heat appliances. Moreover, as some energy providers experiment with variable tariffs and customers invest in home energy management systems that can optimize for market prices in addition to self-consumption, demand profiles will incorporate market feedback that will alter electricity demand profiles further. Future research, thus, has to include novel predictors, like short term price elasticity or technical aspects like the common ripple control in some countries, which, because of the control, could improve the prediction quality. 

\begin{acks}
We thank the \grantsponsor{1}{Bavarian Ministry of Economic Affairs, Regional Development and Energy}{https://www.iuk-bayern.de/} for their financial support of the project "DigiSWM" (\grantnum{1}{DIK-2103-0014}), as part of which the study was carried out.
\end{acks}

\bibliographystyle{ACM-Reference-Format}
\bibliography{References}

\end{document}